\documentclass{article}
\usepackage[utf8]{inputenc}
\usepackage{biblatex}
\usepackage{listings}
\usepackage{color}
\usepackage{hyperref}
\usepackage{graphicx}
\usepackage{caption}
\usepackage{subcaption}
\usepackage{booktabs}
\usepackage{authblk}
\usepackage{geometry}
\usepackage[printwatermark]{xwatermark}
\usepackage{xcolor}
\usepackage{todonotes}

\definecolor{dkgreen}{rgb}{0,0.6,0}
\definecolor{gray}{rgb}{0.5,0.5,0.5}
\definecolor{mauve}{rgb}{0.58,0,0.82}

\title{A Fortran-Keras Deep Learning Bridge for Scientific Computing}

\author[1]{Jordan Ott}
\author[2]{Mike Pritchard}
\author[3]{Natalie Best}
\author[3]{Erik Linstead}
\author[4]{Milan Curcic}
\author[1]{Pierre Baldi}

\affil[1]{Department of Computer Science \protect\\ University of California, Irvine}
\affil[2]{Department of Earth System Science\protect\\ University of California, Irvine}
\affil[3]{Fowler School of Engineering \protect\\ Chapman University}
\affil[4]{Department of Ocean Sciences\protect\\ University of Miami}

\date{}

\bibliography{sources}

\begin{document}

\maketitle
\begin{abstract}
    Implementing artificial neural networks is commonly achieved via high-level programming languages like Python and easy-to-use deep learning libraries like Keras. These software libraries come pre-loaded with a variety of network architectures, provide autodifferentiation, and support GPUs for fast and efficient computation. As a result, a deep learning practitioner will favor training a neural network model in Python, where these tools are readily available. However, many large-scale scientific computation projects are written in Fortran, making it difficult to integrate with modern deep learning methods. To alleviate this problem, we introduce a software library, the Fortran-Keras Bridge (FKB). This two-way bridge connects environments where deep learning resources are plentiful, with those where they are scarce. The paper describes several unique features offered by FKB, such as customizable layers, loss functions, and network ensembles. 
    
    The paper concludes with a case study that applies FKB to address open questions about the robustness of an experimental approach to global climate simulation, in which subgrid physics are outsourced to deep neural network emulators. In this context, FKB enables a hyperparameter search of one hundred plus candidate models of subgrid cloud and radiation physics, initially implemented in Keras, to be transferred and used in Fortran. Such a process allows the model's emergent behavior to be assessed, i.e. when fit imperfections are coupled to explicit planetary-scale fluid dynamics. The results reveal a previously unrecognized strong relationship between offline validation error and online performance, in which the choice of optimizer proves unexpectedly critical. This in turn reveals many new neural network architectures that produce considerable improvements in climate model stability including some with reduced error, for an especially challenging training dataset.
\end{abstract}

\section{Introduction}
The Fortran programming language was originally developed in the 1950s and published in 1957.
It was created to help programmers implement solutions for scientific and engineering problems on the IBM 704 computer, which at the time needed to be written in machine or assembly language. Fortran has been regarded as revolutionary and possibly one of the most influential software products in history \cite{fortranhistory}. Having evolved many times since its creation, with the most recent release in 2018, each version adds new features and capabilities. Fortran initially gained popularity and remains a widely used language due to its fast and efficient computational ability. Additionally, Fortran's strength is its backward compatibility, which allows modern compilers to build code written in the 60s and 70s.

Though not as popular as it once was, Fortran is still used in specialized fields, including oceanography, solid mechanics, computational physics, earthquake simulation, climate modeling, and aerospace. Because of Fortran's continued use, a great deal of legacy code and new code exists.
Unfortunately, it is difficult to rewrite all existing code bases in more mainstream languages, due to their size and complexity. Therefore, when algorithms and extensive libraries are created in modern languages, backwards compatible methods must be developed to make them available in older legacy code, like Fortran.

In recent years, the rise of machine learning and deep learning has led to successful applications in various domains. Substantial improvements in the size of the training sets and available computing power have led to a new wave of implementations \cite{krizhevsky2012imagenet, schmidhuber2015deep}. In turn, this success has increased the usage and dissemination of deep learning. These methods have been applied to a variety of domains, e.g., ranging from remote sensing \cite{zhu2017deep, lahaye2019multi} to computer vision \cite{ott2018deep, tompson2014real, ott2018learning, ott2019exploring, baldicolonoscopy2018}, and to games \cite{agostinelli2019solving, silver2016mastering}. Specifically, within scientific computing, many advancements have been achieved through the application of neural networks. Neural networks have been augmented with physically informed capabilities \cite{raissi2019physics, beucler2020enforcing}, better suiting them for conservation restrictions. Learning partial differential equations \cite{bar2019learning, rudy2017data} has proved valuable in multiple scientific domains. 

The success and popularity of deep learning have inspired the creation of powerful software libraries written in several modern programming languages. However, Fortran is not among the modern languages that benefit from these deep learning libraries. This absence leaves Fortran programmers with few options to implement deep neural networks. 

The implementation of deep neural networks, in Fortran, may be achieved via two primary pathways. One solution is to rewrite all existing deep learning libraries in Fortran. The second solution is to leverage existing frameworks and bridge available functionalities to Fortran. The former is extremely arduous and time consuming, considering the size and scope of existing deep learning packages and the dizzying pace of their evolution \cite{chollet2015keras, abadi2016tensorflow, paszke2017automatic}. 
The latter approach, which this paper describes, is to allow users to leverage the power of existing frameworks while providing a bridge between paradigms where deep learning resources are plentiful and those where they are scarce. In this way, we can leverage aspects of currently available deep learning software libraries, like Keras \cite{chollet2015keras}, and bring them to large-scale scientific computing packages written in Fortran. To this end, we propose the Fortran-Keras Bridge (FKB) -- A two-way bridge connecting models in Keras with ones available in Fortran. The source code is publicly available and can be found here: \url{https://github.com/scientific-computing/FKB}. We begin by reviewing existing Fortran projects that would benefit from the integration of FKB.

\section{Fortran Projects}
\label{related_works}
FKB can be integrated with many existing large-scale and computationally intensive projects written in Fortran. These projects will benefit from the easy integration of neural network models, which FKB makes possible. 

For example, Fortran is used to do a great deal of work in climate and ocean modeling. For instance, the US-produced Community Earth System Model \cite{hurrell2013community} is written in object-oriented Fortran-90; this is the most widely used climate model in the world. So are the other climate simulation codes used by the US Department of Energy \cite{golaz2019doe} and the National Oceanographic and Atmospheric Administration's Geophysical Fluid Dynamics Laboratory \cite{held2019structure}. Meanwhile, the Nucleus for European Modelling of the Ocean (NEMO) engine is used for studying ocean circulation problems on regional and global scales \cite{NEMO} and making future predictions, is also written in Fortran. The Hybrid Coordinate Ocean Model (HYCOM) \cite{HYCOM}, also used for ocean modeling, extends traditional ocean models to allow for a smooth transition from the deep ocean to coastal regimes. Researchers have also developed models for the modeling of waves and wind stress \cite{donelan12waveswind}. The Weather Research and Forecasting Model (WRF), is arguably the most widely used numerical weather prediction models for regional decision support \cite{powers17wrf}. Since its release in 2000, the number of WRF registrations has grown to over 36,000. WRF produces atmospheric simulations with support for special applications, including air chemistry, hydrology, wildland fires, hurricanes, and regional climate, and is again a Fortran-based model.

Fortran has found continued use in solid mechanics packages for implementing finite element methods. Popular packages such as ANSYS \cite{madenci2015finite}, ABAQUS \cite{borgesson1996abaqus}, and LS-DYNA \cite{murray2007users} are written in Fortran or accept Fortran subroutines. Similarly, in earthquake modeling, the SPECFEM3D \cite{SPECFEM3D} package leverages Fortran for simulations.  

The list goes on. Code Saturne \cite{saturne}, developed by Électricité de France, and NEK5000 \cite{nek5000}, are Fortran open-source computational fluid dynamics packages. Code\_Saturne allows for user customization via Fortran subroutines, which is just one application domain for FKB. NEK5000 is actively used in the Center for Exascale Simulation of Advanced Reactors (CESAR) projects. Fortran has also been continually used for molecular modeling within chemistry and physics. The Chemistry at Harvard Macromolecular Mechanics (CHARMM) Development Project has produced a powerful molecular simulation program in Fortran \cite{charmm}. This simulation program primarily targets biological systems but can also be used for inorganic materials. A similar tool, NWChem, has been developed by the Molecular Sciences Software Group at the Pacific Northwest National Laboratory \cite{valiev10NWChem}. NWChem is a computational chemistry software that includes quantum chemical and molecular dynamics functionalities. Within the molecular physics domain, Fluktuierende Kaskade (FLUKA) is a proprietary tool for calculations of particle transport and interactions with matter \cite{fluka}.

The models mentioned above and projects can leverage the FKB library to leverage neural networks within their codebases. For example, neural networks have proven useful in modeling sea surface temperature cooling for typhoon forecasting \cite{typhoon}. Therefore the integration of FKB with tools like NEMO, HYCOM, or WRF models is a possibility. In a recent study of computational fluid dynamics, Ling et al. solve the Reynolds-averaged Navier-Stokes equations, similar to Code\_Saturne and NEK5000. By implementing deep neural networks, the authors report that the architecture improved prediction accuracy \cite{ling_kurzawski_templeton_2016}. 
Finally, the Fluka tool contains a wide range of molecular physics applications, including dosimetry calculations. Vega-Carrillo et al. have shown neural networks aided in the calculation of neutron doses \cite{dosimetry}. For global climate simulation, there is proof that deep neural networks can offer skillful alternatives to assumption-prone approximations of sub-grid cloud and turbulence physics in the atmosphere \cite{rasp2018deep,brenowitz2018prognostic}. We hope that the FKB library enables Fortran users to expand their research and projects to include neural networks.

Having reviewed several Fortran based projects that can leverage FKB, we now introduce the two sides of this bridge. The following sections will develop the foundations on which to anchor each side of this two-way bridge. We start by introducing the deep learning anchor. 

\begin{figure}
\centering
\begin{subfigure}{.5\textwidth}
  \centering
  \includegraphics[width=\linewidth]{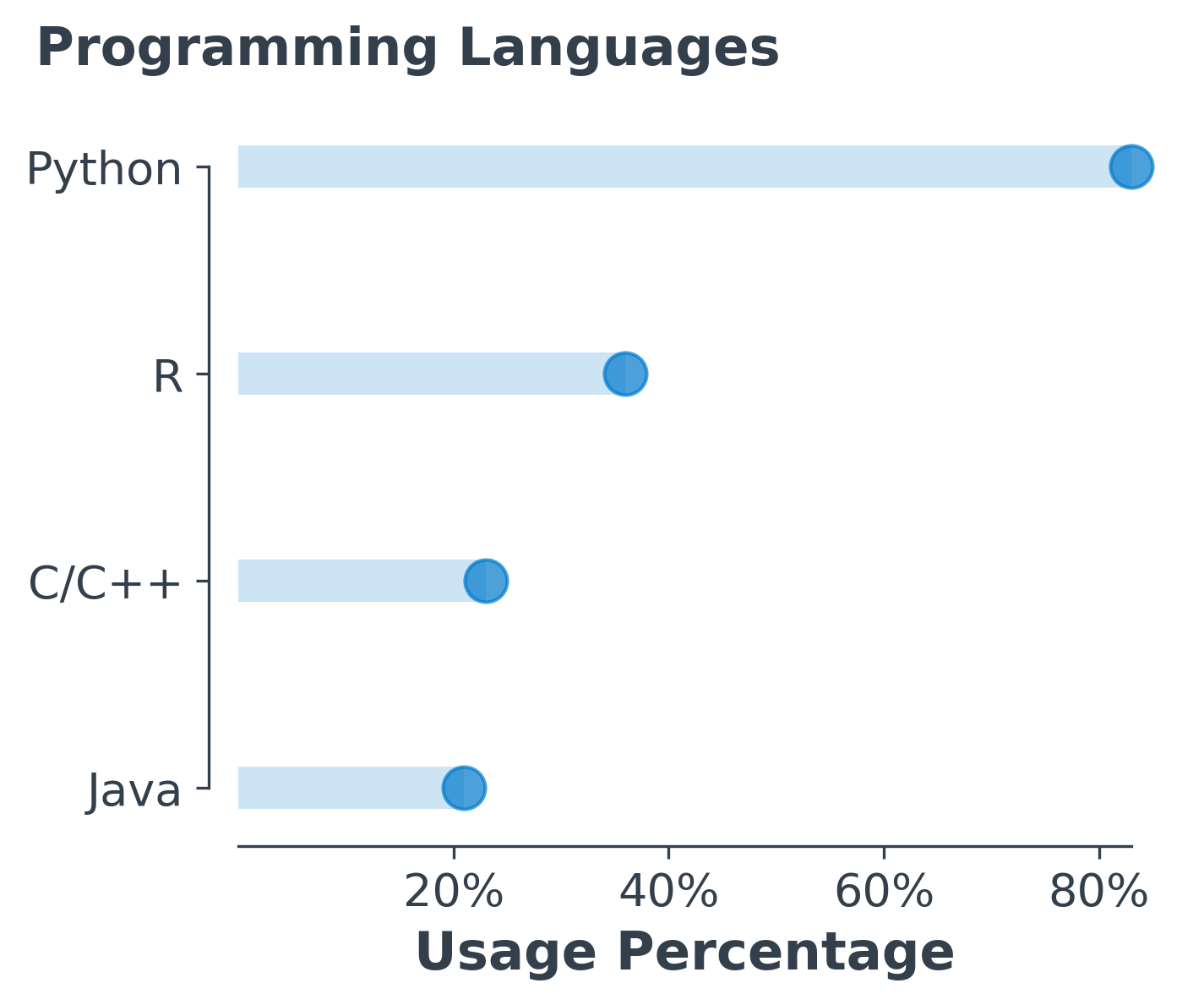}
  \caption{}
  \label{fig:programming_languages}
\end{subfigure}%
\begin{subfigure}{.5\textwidth}
  \centering
  \includegraphics[width=1.1\linewidth]{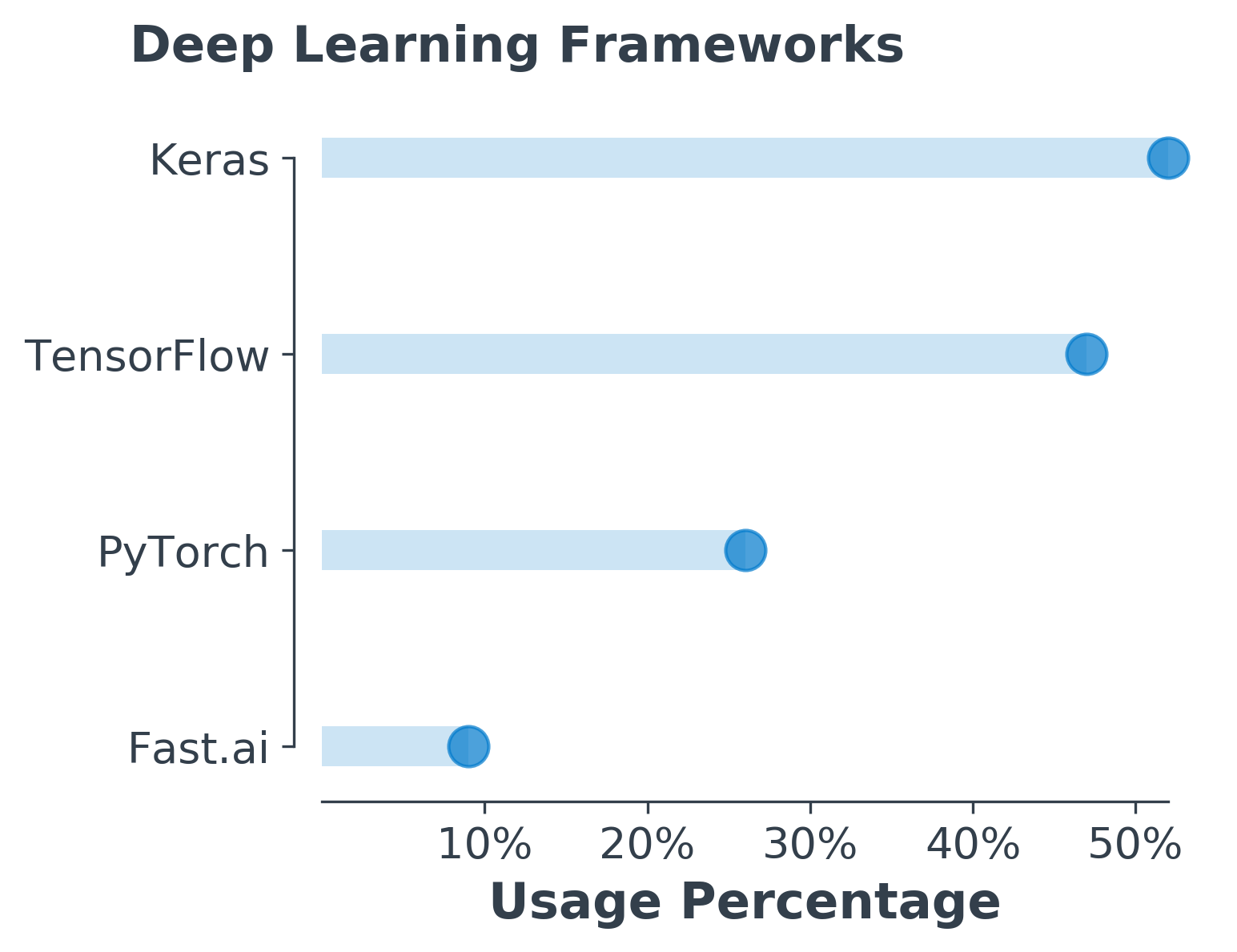}
  \caption{}
  \label{fig:dl_frameworks}
\end{subfigure} 
\caption{(a) Usage of programming languages for machine learning and data science. Statistics are from the 2018 Kaggle ML \& DS Survey \cite{kaggle18}. (b) Usage metrics of deep learning frameworks. Statistics are from the 2019 Kaggle State of Data Science and Machine Learning report \cite{kaggle}.}
\label{fig:plots}
\end{figure}

\begin{figure}
    \centering
    \includegraphics[width=\linewidth]{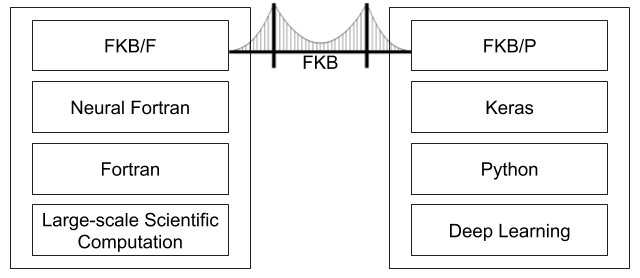}
    \caption{Positioning of FKB within Fortran and Python ecosystems.}
    \label{fig:diagram}
\end{figure}

\section{The Python Anchor (Deep Learning)}

Many programming languages offer tools and libraries for implementing artificial neural networks. However, in recent years, Python has emerged as the clear favorite within this domain. Metrics in Figure \ref{fig:programming_languages} display Python's dominance. Python is used nearly 50\% more than the second most popular language, R. Python's ubiquitous presence in machine learning makes it the obvious choice to leverage existing libraries for Fortran. The question then becomes, which available software library within Python, is best suited to bridge to Fortran?

Of the available deep learning libraries, Keras \cite{chollet2015keras} is the most popular among practitioners (Figure \ref{fig:dl_frameworks}). Keras is an Application Programming Interface (API) built on top of Tensorflow \cite{abadi2016tensorflow}, that provides users the ability to implement quickly, train, and test networks. This convenience encapsulates much of the low-level complexity one must manage when implementing deep networks from scratch. Keras abstracts many of the complicated aspects of Tensorflow while still providing customizability and ease of use. This combination makes Keras the first choice of many for deep learning applications. As a result of its popularity and ease of use, Keras is the clear choice on which to build one end of the two-way bridge. 

Figure \ref{fig:diagram}, depicts the positioning of the Python anchor, FKB/P, within the deep learning ecosystem. The Keras API leverages Python to build deep neural networks. FKB/P resides on top of Keras to access models produced from Keras and transmit them to the Fortran anchor, FKB/F. This structure allows for integration with Fortran applications that wish to leverage deep neural network architectures. Having described the deep learning anchor within Python, the next section develops the foundation for anchoring the bridge with Fortran. 

\section{The Fortran Anchor (Scientific Computing)}
Several attempts have been made to implement neural networks in Fortran, with some success \cite{curcic2019parallel, NEURBT, SAGRAD, Brierley, FANN}. However, many implementations resort to hacking a single-use neural network by hand, or binding code from other languages \cite{FANN}. Along these lines, one may consider accessing Python functionality directly from Fortran, by running a Python instance within Fortran. While providing flexibility and ease of use, this is vulnerable to extreme deficiencies in speed and computational resources. As a result, this solution becomes untenable for large-scale computation projects like the ones described in Section \ref{related_works}.

There are a small number of existing neural network libraries in Fortran \cite{FANN, lary2004using, curcic2019parallel}. The most recent and well developed library is Neural Fortran \cite{curcic2019parallel}, a lightweight neural network library, written natively in Fortran. The Neural Fortran library provides the ability to implement artificial neural networks of arbitrary size with data-based parallelism. Additionally, in benchmark studies, Neural Fortran was shown to have comparable compute performance with Keras while maintaining a lower memory footprint. This library offers a foundation to anchor the Fortran side of the two-way bridge, FKB/F. By extending - and building on top of - Neural Fortran, we can convert Keras models to ones readily available in Fortran and implement them in existing Fortran projects. 

The positioning of FKB within the scientific computing ecosystem is shown in Figure \ref{fig:diagram}. The Fortran anchor, FKB/F, can use models originally constructed and trained in Keras, which can then be transferred to Fortran via FKB/P. To use these models, the Fortran side of FKB implements a neural network library. This portion of FKB can be used within large-scale scientific computation software, like the projects identified in Section \ref{related_works}. 

By leveraging FKB, it becomes seamless to train networks in Python and transfer them to Fortran, to run inside large scale simulations. Similarly, neural network models constructed in Fortran can be transferred to Python for additional analysis, expansion, and optimization - including hyperparameter searches using available tools in Python \cite{hertel2020, snoek2012practical, bergstra2013hyperopt}. As both sides of the bridge have been properly introduced, the following section will describe the specific features and functionalities of FKB.

\section{Features of FKB}
\label{components}

Once a neural network is trained in high-level APIs like Keras, the practitioner has few practical avenues for using this model in Fortran-based projects. One approach may be to hard code network operations inside Fortran while manually moving parameters from the Keras model. Several examples of this can been seen in climate modeling \cite{rasp2018deep,brenowitz2018prognostic,gagne2019machine, gagne2020emulation}.

To provide one specific example, in \cite{rasp2018deep}, the authors trained a deep neural network (DNN) to represent sub-grid cloud and convective energy transport processes, in Keras. To assess its credibility, they needed to test the DNN's two-way interactions when thousands of replicates of it were embedded within a coarse-resolution global atmospheric model, written in Fortran -- neural network emulated clouds interacting with determinstic physical calculations of planetary geophysical fluid dynamics. As the global atmospheric simulator does not offer native neural network support, the authors hardcoded their DNN model into the global simulation software framework. This approach has obvious disadvantages. Every minor change made to the model in Keras requires rewriting the Fortran code. If one wishes to test a suite of models in Fortran, this approach becomes untenable. 
As each network may require different hyperparameters and, as a result, necessitates rewriting and compiling the Fortran code for every new model. This process drastically limits the breadth of available models to be tested within the simulator. This bottleneck is currently a significant roadblock to ongoing debates in the climate simulation community, more broadly, about whether or not to use DNN representations of subgrid physics in next-generation climate modeling. Insufficient testing of diverse candidate neural networks (NN) means that little is known about how minor imperfections in the fit of one NN can amplify when the NN is coupled to fluid dynamics, which is just beginning to be explored \cite{brenowitz2020interpreting}.

These issues demand a solution, in the form of a bridge between Keras and Fortran. The FKB software solves these issues via two key elements. First, it provides a neural network library implemented in Fortran (FKB/F). Second, it offers the ability to parse existing Keras models into formats consistent with the Fortran neural network library (FKB/P). As a result, users can switch, seamlessly, back and forth between Python and Fortran. This context provides a way for iterative neural network tuning (Python) and testing (Fortran), with a simple way to translate between the two software environments. Additionally, FKB offers currently unavailable Fortran specific features for neural networks. It will be useful to highlight those new features while documenting the format to which FKB adheres. The following subsections describe the Python and Fortran anchors' features, FKB/P and FKB/F, respectively.

\subsection{FKB/P}
\label{bridge}
Keras models - once built, trained, and saved - are stored in Hierarchical Data Format 5 (HDF5) files. These files contain the network architecture, weights, biases, and additional information - optimizers, learning rates, gradients, etc. From the HDF5 file, FKB/P parses the network architecture, extracting the number of layers, activation functions, nodes per layer, and all weights and biases. This information is converted to match the Fortran neural network configuration in FKB/F. This allows users to build an equivalent network in Fortran, which can easily be loaded and used within a Fortran environment. If any modifications to the model are made inside Fortran, FKB/P will parse this back into the equivalent HDF5 file to be used in Keras once again. 

On the other hand, networks may be initially constructed in Fortran. After initial training and testing, a user can switch to Keras for further evaluation. From Keras, users can conduct additional testing or hyperparameter tuning where these tools are readily available \cite{hertel2020}. 

The ability to seamlessly pass neural network architectures between Python and Fortran is essential for any practitioner working in this space. This bridge allows users to take advantage of the high-level Keras API - training on computationally efficient GPUs - then to insert their trained model into a Fortran codebase. The functionality provided bridges the chasm between Keras and Fortran.

\subsection{FKB/F}
The Fortran anchor of FKB leverages and extends the original Neural Fortran library. Below we introduce newly implemented features to make Neural Fortran more flexible and able to communicate on the two-way bridge. 

\subsubsection{Custom Layers}
\label{layers}
To implement neural networks in Fortran, FKB leverages and extends the Neural Fortran library \cite{curcic2019parallel}. The prototype Neural Fortran library format that we build on was only capable of implementing a fully connected layer. Forward and backward operations occurred outside this layer - in the network module. An example of this is shown in Listing \ref{ls:old_network_mod}. From the listing, one can observe hard-coded matrix multiplication of layer weights, the addition of biases, and the activation functions inside the network module. This network-level subroutine accesses and modifies individual layer attributes. This rigid format is inconsistent with modern neural network implementation paradigms \cite{chollet2015keras, abadi2016tensorflow, paszke2017automatic}, but it makes it impossible to implement other layers or custom operations. To increase the library's flexibility, operations must be encapsulated inside the layer, consistent with current practice.  

\begin{lstlisting}[caption={Original code from \cite{curcic2019parallel}. Layer operations occur inside the network module, limiting flexibility.},captionpos=b,label={ls:old_network_mod}]
pure subroutine fwdprop(self, x)
    ! Performs the forward propagation and stores arguments to activation
    ! functions and activations themselves for use in backprop.
    class(network_type), intent(in out) :: self
    real(rk), intent(in) :: x(:)
    integer(ik) :: n
    associate(layers => self % layers)
        layers(1) % a = x
        do n = 2, size(layers)
            layers(n) % z = matmul(transpose(layers(n-1) % w), layers(n-1) % a) + layers(n) % b
            layers(n) % a = self % layers(n) % activation(layers(n) % z)
        end do
    end associate
end subroutine fwdprop
\end{lstlisting}

In FKB we introduce an extendable layer type module (Listing \ref{ls:new_forward}). To implement a layer, one simply extends the layer type and specifies the construction of the forward and backward functions. Adhering to this format offers several advantages. By restructuring the format of the library, we offer the ability to implement arbitrary layers. Additionally, in the network module, all layers are stored in an array of pointers. This leads to the encapsulated version shown in Listing \ref{ls:new_forward} wherein a forward pass, in the network module, calls the layer-specific forward function. In this way, all operations are confined to the layer module, and the output from one layer is passed as input to the next. 
      
\begin{lstlisting}[caption={Forward pass in the FKB network module. Each layer simply calls its own forward function. The technical operations occur within each layer.},captionpos=b,label={ls:new_forward}]
function output(self, input) result(last_layer_output)
    ...
    ! iterate through layers passing activation forward
    do n = 1, size(layers)
        call layers(n) % p % forward(layers(n-1) % p % o)
    end do
    ! get output from last layer
    last_layer_output = layers(size(layers)) % p % o
end function output
\end{lstlisting}

FKB supports fully connected or dense layers, dropout \cite{srivastava2014dropout, baldi2014dropout}, and batch normalization \cite{ioffe2015batch}. Shown in Listing \ref{batch_norm} is an example of extending the layer\_type to implement a Batch Normalization layer. This format translates to increased functionality and customizability to the user. As a result, more standard layers from Keras are available, while giving users the flexibility to implement their own custom operations.

\begin{lstlisting}[caption={Example of extending the layer\_type to implement Batch Normalization},captionpos=b,label={batch_norm}]
! BatchNorm layer - extends from base layer_type
!   Implements batch normalization
type, extends(layer_type) :: BatchNorm
    ! epsilon parameter
    real(rk) :: epsilon
contains
    procedure, public, pass(self) :: forward => batchnorm_forward
    procedure, public, pass(self) :: backward => batchnorm_backward
end type BatchNorm
\end{lstlisting}

\subsubsection{Training in Fortran}
It is necessary to distinguish between the terms \textit{offline} versus \textit{online} for the following section. These terms serve to distinguish two different settings in which a neural network can be used in a Fortran computing package.
Both settings can make use of historical or simulated data to train an artificial network. The distinguishing feature is how the predictions of a model are used. In an online setting, predictions from the model are used to evolve a physical process. 
The predictions at one time step effect how the system acts at the following time step. As a result, inputs to the model will change based on how the model acted in the past. In offline settings, this is not the case. Predictions made in the past do not affect the input to the model in the future. 


In many cases, offline training may be sufficient to learn a model, if enough prior data is available. However, in some cases, online training may be the method of choice. To this end, FKB is equipped to handle backpropagation for gradient descent optimization of a specified cost function.    

The layer encapsulation mentioned above of forward and backward operations (Section \ref{layers}) becomes extremely valuable in training. Instead of all computations occurring within the network module \cite{curcic2019parallel}, they are contained in layer-specific functions. Much like the forward pass, backward operations occur in the layer. In this fashion, each layer is responsible for computing its gradients with respect to its parameters and returning the gradient with respect to the layer below it. 

Online training can serve a variety of purposes. First, a neural network model may be learned entirely in Fortran, based on the evolving state variables during the integration of a physical dynamical system simulation, and then transferred to Keras after the fact. In this setting, the ground truth, from the simulator, is passed to the network for it to calculate its errors and update its parameters accordingly through backpropagation. Second, online training could serve to provide gentle corrections to an imperfect pretrained model, for instance, to hedge against the amplification of its imperfections that are only revealed once the NN is coupled to other physical calculations. Here a model is trained offline in Keras and transferred to Fortran (Section \ref{bridge}). In some cases, for a variety of reasons, the offline training data may have a differing distribution than that of the online data. In such a setting, it proves beneficial to offer slight corrections to the network. Finally, a secondary model may be constructed to learn and compensate for the deficiencies in the primary model. In this way, the two networks work together to balance out any instability issues. 

The ease of use and proper format directly results from the encapsulation of layer operations. Online training offers a solution to tackle a suite of potential problems. As a result, models may be updated with slight corrections or learned entirely online.

\subsubsection{Custom Loss Functions}
\label{loss_functions}
In many applications, practitioners may wish to optimize a unique quantity - a function other than a mean squared error or cross-entropy. This is common when target variables interact or additional information is known about their relationship in a desired application. For example, in modeling any physical system, predictions from a neural network must not violate physical constraints - energy cannot be created or destroyed in the system. To satisfy this restriction, a loss function can be written to quantify the amount of violation of physical properties. This construction can then be minimized to alleviate constraint infractions \cite{beucler2020enforcing}.

The implementation of custom loss functions is standard for high-level APIs like Keras, Tensorflow, and PyTorch to provide this ability in their codebase \cite{chollet2015keras, abadi2016tensorflow, paszke2017automatic}. As FKB is designed for those working in the physical sciences where environmental, physical, or application-specific constraints are common, it provides the ability to implement custom loss functions. 
To take advantage of this functionality, users must implement their desired loss function, just as they would in Keras. As FKB does not provide automatic differentiation, the derivatives with respect to the input are also required for training. Once these functions have been specified they can be dropped into the existing framework and run normally, much like Keras.  

\begin{lstlisting}[caption={Implementation of crossentropy loss function and the corresponding derivation with respect to the input logits.},captionpos=b,label={ls:crossentropy}]
real(rk) function crossentropy_loss(self, y_true, y_pred)
    ! Given predicted and expected output, returns the scalar loss 
    class(network_type), intent(in out) :: self
    real(rk), intent(in) :: y_true(:), y_pred(:)
    
    loss = - sum(y_true * log(y_pred))
end function loss

function d_crossentropy_loss(self, y_true, y_pred) result(loss)
    ! Given predicted and expected output 
    ! returns the loss with respect to softmax input
    class(network_type), intent(in out) :: self
    real(rk), intent(in) :: y_true(:), y_pred(:)
    real(rk), allocatable :: loss(:)
    
    loss = y_pred - y_true
end function d_loss
\end{lstlisting}






This capability is demonstrated through the implementation of the cross-entropy loss function in Listing \ref{ls:crossentropy}. To implement this previously unavailable loss function, we first declare two functions. First, the cross-entropy scalar loss is. Second, the loss with respect to the input logits is derived. These two functions are then referenced as the loss and d\_loss, respectively. By providing this functionality, users may leverage a variety of loss functions that can be used to minimize application-specific quantities. Once described, they may be included with the existing framework and used during online training.  

\subsubsection{Ensembles}
Ensembles consist of different models, each trained on the same, or bootstrapped, data. The output of the ensemble will be an average of all its member's predictions. In machine learning, ensembles of models typically perform better than any one of its members alone. The ensemble strategy exploits the fact that each model will make different errors. Therefore, when averaged together, these predictions become more accurate, as certain errors get smoothed out. A consensus from machine learning practitioners is ensembling gives 1-2\% improvement in performance \cite{chollet2018deep}. 

As a result of this averaging, ensembles provide a boost in performance as well as additional robustness. In domains where physical constraint violations yield stability issues, ensembles may be applied to dampen these problems. By averaging across many networks, the instability of any one model will be drastically reduced in the presence of more sound predictions.

The functionality provided requires the user to specify a directory that contains the models of interest and a desired amount of noise. The ensemble type will read in each model and construct a network corresponding to each of them. To get a prediction from the ensemble, an input vector is passed to it. For non-zero amounts of noise, Gaussian noise is applied to the input vector each time it is passed to an ensemble member. This allows each member to see a slightly different variant of the input, increasing the robustness of prediction around that point. This operation runs in parallel using OpenMP, where each network can be given its thread to expedite computation; such an approach could easily be adapted via OpenACC for GPU-based threading of large ensemble network calculations. Following the computation, the predictions are averaged together, and the final output is given. 
  
\section{Case Study}
The following section provides a case study demonstrating an application of FKB to experimental next-generation climate modeling. The Superparameterized Community Atmospheric Model version 3.0 (SPCAM3) is used for all simulations in this study. SuperParameterization is an approach that confronts the decades-long problem of representing subgrid cloud physics in climate models by embedding thousands of limited-domain explicit sub-models of moist convection within a conventional planetary-scale model of the large scale atmosphere \cite{grabowski2001,khairoutdinov2005simulations, khairoutdinov2008evaluation, thayer2009role}. This approach tends to involve two orders of magnitude more computational intensity per unit area of the simulated earth, but recently Rasp et al. used a deep neural network to emulate all of the expensive subgrid cloud resolving models' (CRM) influence on the planetary host at drastically reduced computational expense \cite{rasp2018deep}. This study, along with others in the emerging climate modeling literature \cite{brenowitz2018prognostic} have demonstrated the potential advantages of a data-driven approach for addressing the critical unresolved effects of clouds and convection on planetary climate, as compared to previous, heuristic based, approximations to subgrid physics. However, the idea of emulating turbulence in climate simulation is still an emerging one, with unclear trade-offs, including frequent instabilities when NN emulators are coupled with fluid dynamics, which the community is seeking to learn how to control \cite{brenowitz2018prognostic}. It has even been questioned whether the offline skill of such emulators, during their training, is predictive of their online performance \cite{rasp2019online,gagnelorenz20}, an important open question.

These questions are understudied primarily due to the lack of the simple software interface that FKB now enables for climate scientists to test diverse candidate neural networks, and ensembles within planetary climate models.

To illustrate an advance on this front we now apply FKB to shed new light on two related questions currently in debate:
\begin{enumerate}
    \item Does offline performance translate to online model performance \cite{rasp2019online,gagnelorenz20}?
    \item Which neural network hyperparameters most affect online performance? 
\end{enumerate}

Using FKB, the study can be broken into two stages. First, a suite of 108 candidate neural network models of convection are trained, via Keras, on simulated data from the SPCAM3. Second, the models are converted to Fortran and run online (i.e. coupled to planetary fluid dynamics) in the SPCAM3 simulator. The number of steps serves as a preliminary metric of performance until catastrophic failure. It is clear that in the absence of the FKB library, running hundreds of candidate neural network submodels of convection within the Fortran based model of the rest of the planet's atmosphere would be nearly impossible. As each network contains various hyperparameters, each with different weights and biases learned during training, including layer-specific properties such as optional use of dropout or batch-normalization. To leverage the FKB library with SPCAM3, we simply compile the neural network library in advance and link it to the compilation of SPCAM3. Documentation steps for the implementation of this case study are provided here: \url{https://github.com/scientific-computing/FKB/blob/master/SPCAM_Instructions.md}.

The input to this neural network model is a 94-dimensional vector. Features include vertically resolved vectors representing the large scale (host model) temperature, humidity, meridional wind vertical structure, surface pressure, incoming solar radiation, sensible heat flux, and latent heat flux scalars. The output of the network is a 65-dimensional vector composed of the embedded models' influence on their host - i.e. the sum of the CRM and radiative heating rates, the CRM moistening rate, the net radiative fluxes at the top of the atmosphere and surface of the earth, and the precipitation. 

The training data used here are challenging to fit, as they come from an enhanced version of the CRM training data that was originally studied by \cite{rasp2018deep}. In superparameterized simulations, one can control the degrees of freedom of the interior resolved scale through the room available for interesting forms of sub-grid storm organization to form. One can control the physical extent (i.e. number of columns used in) each embedded CRM array \cite{pritchard2014restricting}.  In \cite{rasp2018deep}, CRM arrays with only 8 columns (32-km extent, given the 4-km horizontal resolution) were used. Here we quadruple the extent (from 32 km to 128 km, i.e. from 8-columns to 32-columns) to improve its physical realism. Despite several attempts, these data have never been fit successfully. NNs trained from the enriched data tend to produce crashes within just a few simulated weeks after they are embedded in the climate model (see discussion of ``NN-unstable'' by \cite{brenowitz2020interpreting} for details). 

\begin{table}[]
\footnotesize
\centering
    \begin{tabular}{@{}ll@{}ll@{}}
    \toprule
    Name                        & Options    & Parameter Type                   \\ \midrule
    Batch Normalization         & {[}yes, no{]}             &   Choice         \\
    Dropout                     & {[}0, 0.25{]}             &   Continuous      \\
    Leaky ReLU coefficient      & {[}0 - 0.4{]}             &   Continuous      \\
    Learning Rate               & {[}0.00001 - 0.01{]}      &   Continuous (log) \\
    Nodes per Layer             & {[}128,256,512{]}         &   Discrete        \\
    Number of layers            & {[}4 - 11{]}              &   Discrete         \\
    Optimizer                   & {[}Adam, RMSProp, SGD {]} &   Choice           \\
    \bottomrule
    \end{tabular}
\caption{Hyperparameter Space}
\label{hyperparameter-table}
\end{table}

Our working hypothesis is that historical failures in free-running tests when emulators are trained on higher quality CRM training data reflect a broader issue of insufficient hyperparameter tuning in climate model applications. To address this, we conducted neural network optimization via a random search using SHERPA \cite{hertel2020}, a Python library for hyperparameter tuning. We detail the hyperparameters of interest in Table \ref{hyperparameter-table}, as well as the range of available options during the search. The hyperparameters of interest consisted of whether or not to use batch normalization, the amount of dropout, the leaky ReLU coefficient, learning rate, nodes per layer, the number of layers, and the optimizer. The random search algorithm has the advantage of making no assumptions about the structure of the hyperparameter search problem and is ideal for exploring a variety of settings. 

We attained 108 candidate neural network model configurations, each trained for 25 epochs with early stopping monitoring the validation loss. Following the offline training stage, the neural network models were converted into their Fortran counterparts and ran inside SPCAM3. We underscore that this critical step would have been prohibitive using standard tools that have required manual translation of each candidate model. However, by leveraging the FKB library, each model was loaded independently into Fortran and run as the subgrid physics emulator inside SPCAM3's host planetary model, of the large-scale atmospheric state. Each model was coupled to fluid dynamics, to run a wide ensemble of prognostic tests across an unprecedented diversity of candidate neural network architectures. Each of the one hundred and eight candidate neural network models - with their various numbers of layers, layer-specific settings (batch-normalization, relu magnitude, etc), nodes per layer, weights, and biases - were run online, all without rewriting any Fortran code. 

In order to address the first question and evaluate a neural network model's performance, we compare its validation MSE during training with the time-to-failure of the online tests in which 8,192 instances of the NN, spaced at regular intervals around the globe, are coupled interactively to their host global atmospheric model of large scale geophysical fluid dynamics. This yields Figure \ref{fig:results}, which sheds new light on the offline vs. online relationship. 

The results in this figure demonstrate a relationship between offline validation error and online performance. There is a distinct, negative, relationship between offline MSE and online stability (Spearman correlation of $-0.73$; $p= 4.961e^{-19}$. Intriguingly, the mean-squared error loss of our multi-layer perceptron is a reasonable predictor of stability once coupled to the climate model, insofar as the time-to-failure is concerned. This finding is interesting in the context of the recent speculation by \cite{rasp2019online} that such a relationship might not exist using similar NNs in a similar setting, as well as the comments by \cite{gagnelorenz20} about similar incongruities even in reduced-order dynamical systems when emulated with GANs. 

\begin{figure}
    \centering
    \includegraphics[width=\linewidth]{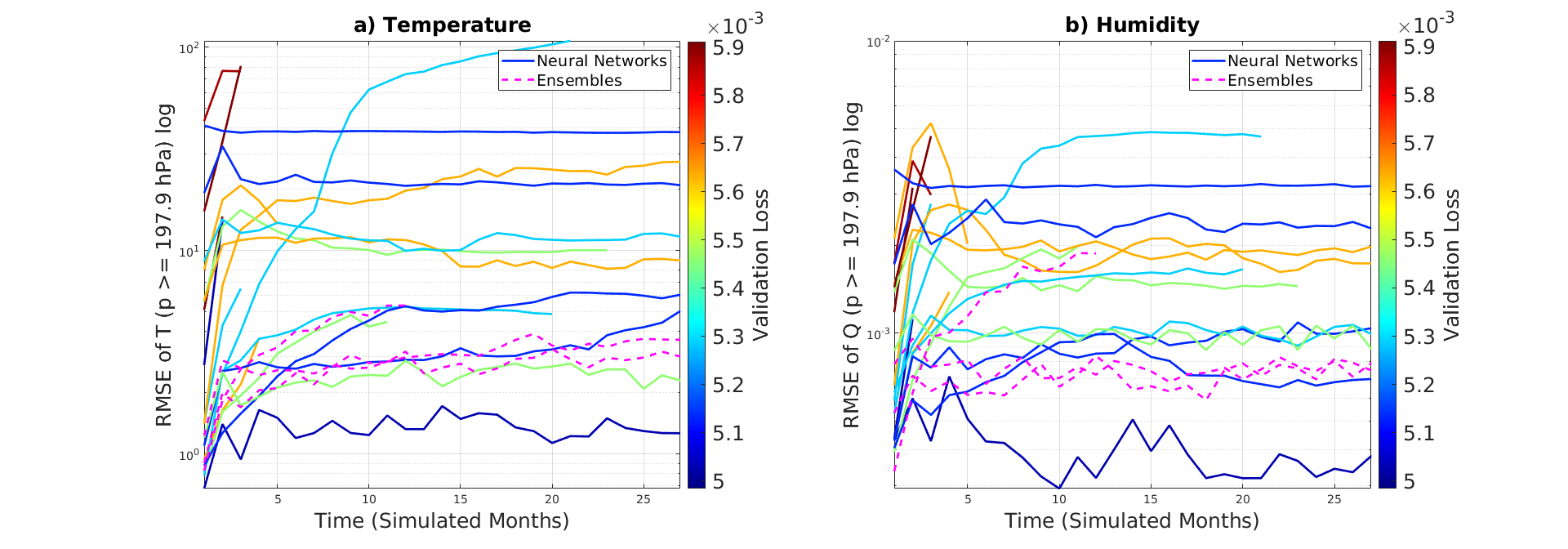}
    \caption{The time-evolution of the tropospheric (a) temperature and (b) humidity biases, colorized by the offline validation error}
    \label{fig:t_q}
\end{figure}

\begin{figure}
\centering
\begin{subfigure}{.33\textwidth}
  \centering
  \includegraphics[width=\linewidth]{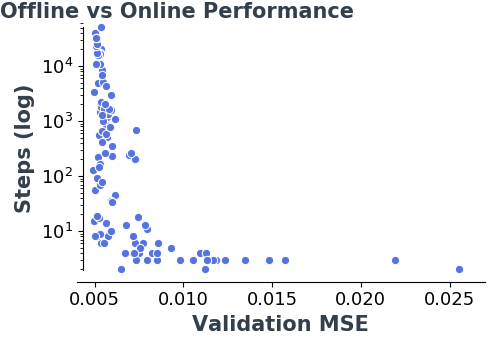}
  \caption{}
  \label{fig:results}
\end{subfigure}%
\begin{subfigure}{.33\textwidth}
  \centering
  \includegraphics[width=\linewidth]{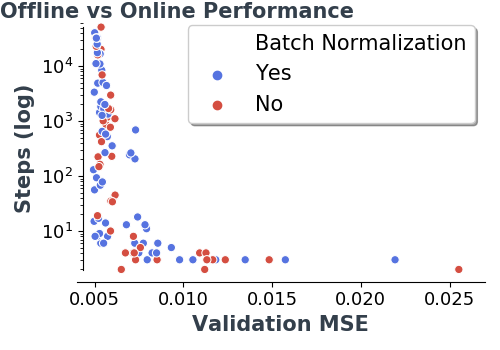}
  \caption{}
  \label{fig:batch_norm}
\end{subfigure} 
\begin{subfigure}{.33\textwidth}
  \centering
  \includegraphics[width=\linewidth]{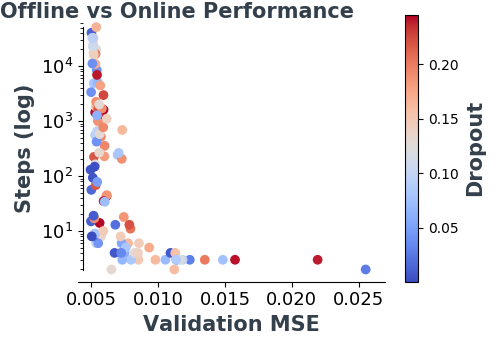}
  \caption{}
  \label{fig:dropout}
\end{subfigure}\\

\begin{subfigure}{.33\textwidth}
  \centering
  \includegraphics[width=\linewidth]{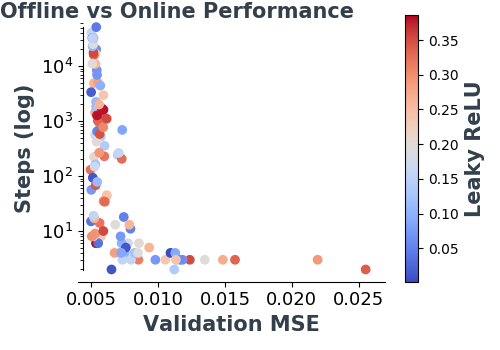}
  \caption{}
  \label{fig:leaky_relu}
\end{subfigure}%
\begin{subfigure}{.33\textwidth}
  \centering
  \includegraphics[width=\linewidth]{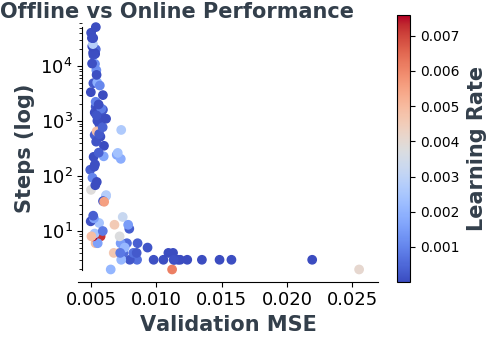}
  \caption{}
  \label{fig:lr}
\end{subfigure} 
\begin{subfigure}{.33\textwidth}
  \centering
  \includegraphics[width=\linewidth]{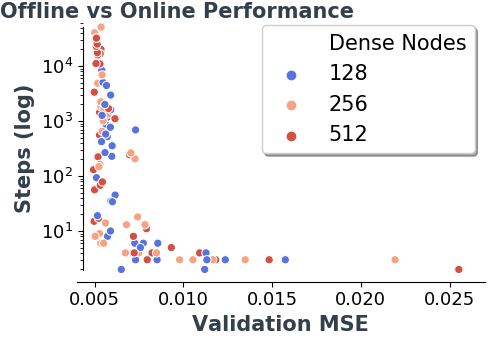}
  \caption{}
  \label{fig:num_dense_nodes}
\end{subfigure} \\

\begin{subfigure}{.33\textwidth}
  \centering
  \includegraphics[width=\linewidth]{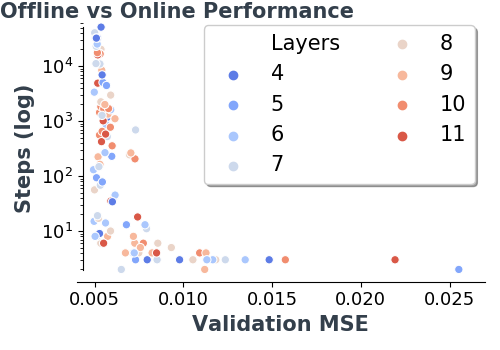}
  \caption{}
  \label{fig:num_layers}
\end{subfigure}%
\begin{subfigure}{.33\textwidth}
  \centering
  \includegraphics[width=\linewidth]{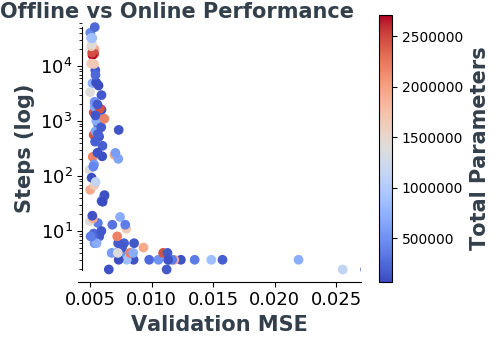}
  \caption{}
  \label{fig:num_params}
\end{subfigure}
\begin{subfigure}{.33\textwidth}
  \centering
  \includegraphics[width=\linewidth]{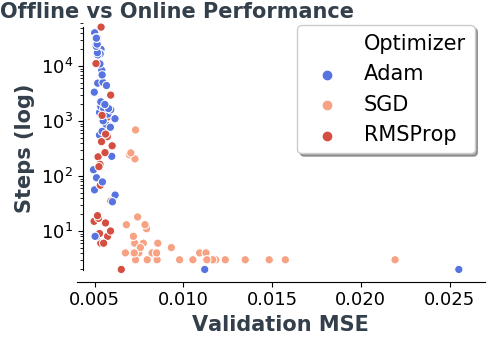}
  \caption{}
  \label{fig:optimizer}
\end{subfigure}
\caption{Offline performance - validation mean squared error (MSE) - vs online performance - number of steps until crash.(a) All models. (b) By batch normalization usage. (c) By Dropout amount. (d) By leaky ReLU coefficient. (e) By learning rate. (f) By number of dense nodes per layer. (g) By number of layers. (h) By total number of model parameters. (i) By optimizer type.}
\label{fig:plots}
\end{figure}

Of course, stability alone is a necessary but not a sufficient condition of prognostic success, which also requires an in-depth analysis of biases in the simulated climate. Figure \ref{fig:t_q} shows the time-evolution of the tropospheric temperature and humidity biases, colorized by the offline validation error. These metrics reveal that although our search has uncovered many runs that are ``stable'' - can run without catastrophically crashing for several months - most of these runs would not be very useful in an operational setting. Almost all NNs exhibit major errors in the simulated climate, having drifted to erroneous attractors with root-mean-square errors in temperature frequently above 10 K. However, the NN that produced the best offline validation error stands out as having the combined desired qualities of stability and skill with temperature biases of less than 2 K, competitive with \cite{rasp2018deep}. Interestingly, coupling instead to the ensemble mean of a few of the best-ranked models (magenta dashed lines) does not outperform coupling to the best fit model, the value of having found it using SHERPA (Figure \ref{fig:t_q}). 

In short, we have produced a successful coupled simulation that was particularly challenging without formal hyper-parameter tuning and FKB. This result suggests that sufficient hyperparameter tuning may be critical to solving chronic instability in climate model applications of DNNs for subgrid physics. 

The second question naturally arises as to which of the hyperparameters are most impactful to the online performance. To assess this, Figure 2b-i decomposes the sensitivity of the baseline relationship to individual hyperparameter choices. The choice of optimizer is shown to correlate most strongly with online performance (Figure \ref{fig:optimizer}). This finding is confirmed by Spearman values, shown in Table \ref{tab:cor}. The optimizer hyperparameter has the largest absolute correlation value with online performance. No other hyperparameter shows as clear a distinction in correlation that is evident in the choice of optimizer, including the network depth and total number of parameters, which are known to be important to offline fits for this problem \cite{gentine2018could}, but are surprisingly not as predictive of coupled skill as the choice of optimizer, whose impact has not previously been isolated (for this application).

Further investigation into the specific optimizer used, reveals the SGD optimizer to perform poorly; NNs fit with SGD never run longer than 1,000 steps when coupled online (Figure \ref{fig:optimizer}). Again the visual intuition from Figure \ref{fig:optimizer} is confirmed by Spearman correlation values. SGD, Adam, and RMSProp have Spearman values of $-0.6670$, $0.5936$, $0.0586$ respectively. These values demonstrate that the use of SGD is negatively correlated with online performance, whereas Adam positively correlates with online performance. This result leads one to speculate that increased improvements in online skill may be realized from more advanced optimizers with enhanced gradient update schedules. 

\begin{table}[]
    \centering
    \begin{tabular}{lrr}
        \toprule
        {} &  Correlation &       P-Value \\
        \midrule
        BatchNorm      &     0.0859 &  3.7896e-01 \\
        Dropout         &     0.1919 &  4.7591e-02 \\
        Leaky ReLU    &     0.0055 &  9.5465e-01 \\
        Learning Rate              &    -0.2087 &  3.0923e-02 \\
        Dense Nodes &     0.1427 &  1.4249e-01 \\
        Layers      &     0.0410 &  6.7491e-01 \\
        Optimizer       &    \textbf{-0.6998} &  5.0177e-17 \\
        Parameters      &     0.1528 &  1.1609e-01 \\
        \bottomrule
    \end{tabular}
    \caption{Spearman correlation of corresponding hyperparameter with online performance, and associated p-value.}
    \label{tab:cor}
\end{table}

Finally, after answering the two questions motivating this case study, we can compare the results of the best performing model with that of previously published models of \cite{rasp2018deep} when applied to the challenging limit of CRMs with 32-km horizontal extent. The model proposed by Rasp et al. was a single deep neural network. The hyperparameter space of this model was not fully explored online in large part due to the laborious process required to transfer those models into Fortran. The Rasp et al. model (provided by the authors) ran for 128 steps before crashing due to instability issues. The five best models achieved in this study ran to completion of a 5-year simulation, i.e. for 87,840 steps; of these, two of the five models further exhibited root-mean-square errors in simulated tropospheric temperature of less than 2 degrees Celsius. This dramatic improvement in stability is a direct result of the ease with which a wide variety of models (identified by SHERPA) can be transferred between Python and Fortran (thanks to FKB). We also note that this method is preferable to another approach that was recently proposed to begin stabilizing the same model, through small-amplitude Gaussian input perturbation \cite{brenowitz2020interpreting} - a strategy that, while promising, adds computational expense and introduces out-of-sample extrapolation issues that can be avoided with the brute-force optimization and wide-ensemble prognostic testing path to stabilization we have outlined here.

This case study has investigated two closely entangled questions: 1) Does offline performance correspond to online model performance? 2) What neural network hyperparameters most effect online performance?  Both of these questions have been answered by leveraging the FKB library. The library offers the ability to expeditiously transfer models trained in Keras to Fortran, where they may be run online in existing simulators. In the absence of FKB, neither one of these questions could be approached without unreasonable human intervention, as the operational target is a climate model with over a hundred thousand lines of code written in Fortran.

\section{Conclusion}

The ubiquitousness of deep learning has resulted from extensive free and open source libraries \cite{chollet2015keras, abadi2016tensorflow, paszke2017automatic}. Deep learning's success and popularity merit its integration in large-scale computing packages, like those written in Fortran. Instead of rewriting all existing libraries in Fortran, we introduced a two-way bridge between low-level, Fortran, and Python through the FKB Library. The library provides researchers the ability to implement neural networks into Fortran code bases while being able to transfer them back and forth with Keras. 

Fortran, which has been a staple within computationally intensive fields for decades, will undoubtedly see continued use due to its fast computational ability and vast amounts of legacy code. The FKB library enables users to access many features of the Keras API directly in Fortran, including the ability to create custom layers and loss functions to suit their needs. We demonstrate the integrability of FKB through our case study involving the SPCAM3 simulator. An advantage of FKB is its ease of use, demonstrated by its ability to be compiled in advance and once linked can be easily leveraged in existing large scale simulators, as we have illustrated for the application of multi-scale physical simulations of the global atmosphere.

\section{Acknowledgments}
The work of JO and PB is supported by NSF NRT grant 1633631. MP acknowledges NSF funding from OAC-1835863 and AGS-1734164. This research also used HPC resources of the Extreme Science and Engineering Discovery Environment (XSEDE), which is supported by the National Science Foundation under grant number ACI-1548562 \cite{towns_xsede_2014} and allocation number TG-ATM190002.

\section{Conflict of Interest}
The authors declare that there is no conflict of interest regarding the publication of this paper.

\printbibliography
\end{document}